\def\year{2021}\relax
\documentclass[letterpaper]{article} % DO NOT CHANGE THIS
\usepackage{aaai21}  % DO NOT CHANGE THIS
\usepackage{times}  % DO NOT CHANGE THIS
\usepackage{helvet} % DO NOT CHANGE THIS
\usepackage{courier}  % DO NOT CHANGE THIS
\usepackage[hyphens]{url}  % DO NOT CHANGE THIS
\usepackage{graphicx} % DO NOT CHANGE THIS
\usepackage{natbib}  % DO NOT CHANGE THIS AND DO NOT ADD ANY OPTIONS TO IT
\usepackage{caption} % DO NOT CHANGE THIS AND DO NOT ADD ANY OPTIONS TO IT
\RequirePackage[linesnumbered,ruled,vlined]{algorithm2e}
\usepackage{booktabs}
\newcommand{\tabincell}[2]{\begin{tabular}{@{}#1@{}}#2\end{tabular}}
\usepackage{amsfonts}
\usepackage{multirow}
\title{PointManifold: Using Manifold Learning for Point Cloud Classification}
\author{

    Dinghao Yang,
    Wei Gao\thanks{Corresponding author: Wei Gao.}
    \\
}
\affiliations{
    School of Electronic and Computer Engineering, Peking University
    PKU Campus, No. 2199, Lishui Road, Xili Lake, Nanshan District, Shenzhen 518055, China\\
    dinghow@stu.pku.edu.cn, gaowei262@pku.edu.cn

}
\begin{document}

\maketitle

\begin{abstract}
	In this paper, we propose a point cloud classification method based on graph neural network and manifold learning. Different from the conventional point cloud analysis methods, this paper uses manifold learning algorithms to embed point cloud features for better considering the geometric continuity on the surface. Then, the nature of point cloud can be acquired in low dimensional space, and after being concatenated with features in the original three-dimensional (3D) space, both the capability of feature representation and the classification network performance can be improved. We propose two manifold learning modules, where one is based on locally linear embedding algorithm, and the other is a nonlinear projection method based on neural network architecture. Both of them can obtain better performances than the state-of-the-art baseline. Afterwards, the graph model is constructed by using the k nearest neighbors algorithm, where the edge features are effectively aggregated for the implementation of point cloud classification. Experiments show that the proposed point cloud classification methods obtain the mean class accuracy (mA) of 90.2\% and the overall accuracy (oA) of 93.2\%, which reach competitive performances compared with the existing state-of-the-art related methods.
\end{abstract}

\section{Introduction}

%On the one hand, some 2D deep learning task has meet the effect of practical application scenarios, such as image classification and traditional object detection. 
%more and more attention is applied to the 3D data analysis domain \cite{ahmed2018pointcloud}, 
With the developments of laser radar and other imaging instruments, three-dimensional (3D) data is becoming much more easily available. Consequently, the effective processing and analysis methods should be investigated for 3D-related applications. As the representative 3D data, point cloud has been widely adopted in indoor navigation, autopilot, and augmented reality, etc. The effective classification of point clouds can be helpful for the better understanding of intelligent systems to different complicated environments. Therefore, the accurate classification of point clouds plays an important role in related practical applications.

%With the improvement of computing power and the substantial increase of data, deep learning \cite{lecun2015deep} technology has made breakthroughs in recent years, e.g. remarkable achievements in computer vision, natural language processing and other fields. 
%Deep neural network performs much better than traditional methods in diverse tasks, such as image classification, face recognition, object detection, semantic segmentation, et al. 

With the improvement of computing power and the substantial increase of data, deep learning \cite{lecun2015deep} technology has become more and more popular for point cloud classification \cite{maturana2015voxnet, qi2016volumetric, su2015multi}. Particularly, graph neural network \cite{zhang2018graphsurvey} develops rapidly, where many data sources can be effectively represented as graphs for modeling and analysis, such as two-dimensional (2D) image, social networking, 3D point sets. Due to the fact that deep learning has the powerful capability in fitting potential relationship among graph nodes, it can fulfill the complicated modeling tasks for high-dimensional data. For 3D point cloud, the properties of data can be actually suitable for graph representation, and therefore the graph neural networks-based analysis method has emerged as a promising exploration direction \cite{zhang2018PointgCN,wang2019dynamic,li2019deepgcns, xu2020grid}.

%Deep learning has the powerful capability in fitting potential relationship among graph nodes, and by designing specific loss functions, it can have the ability to deal with various types of tasks. The properties of point is suitable for graph representation, and thus using graph neural networks to analyse point cloud data has been a valuable exploration direction.

%further promoting the 

\begin{figure}[t]
\begin{center}
\includegraphics[width=\linewidth]{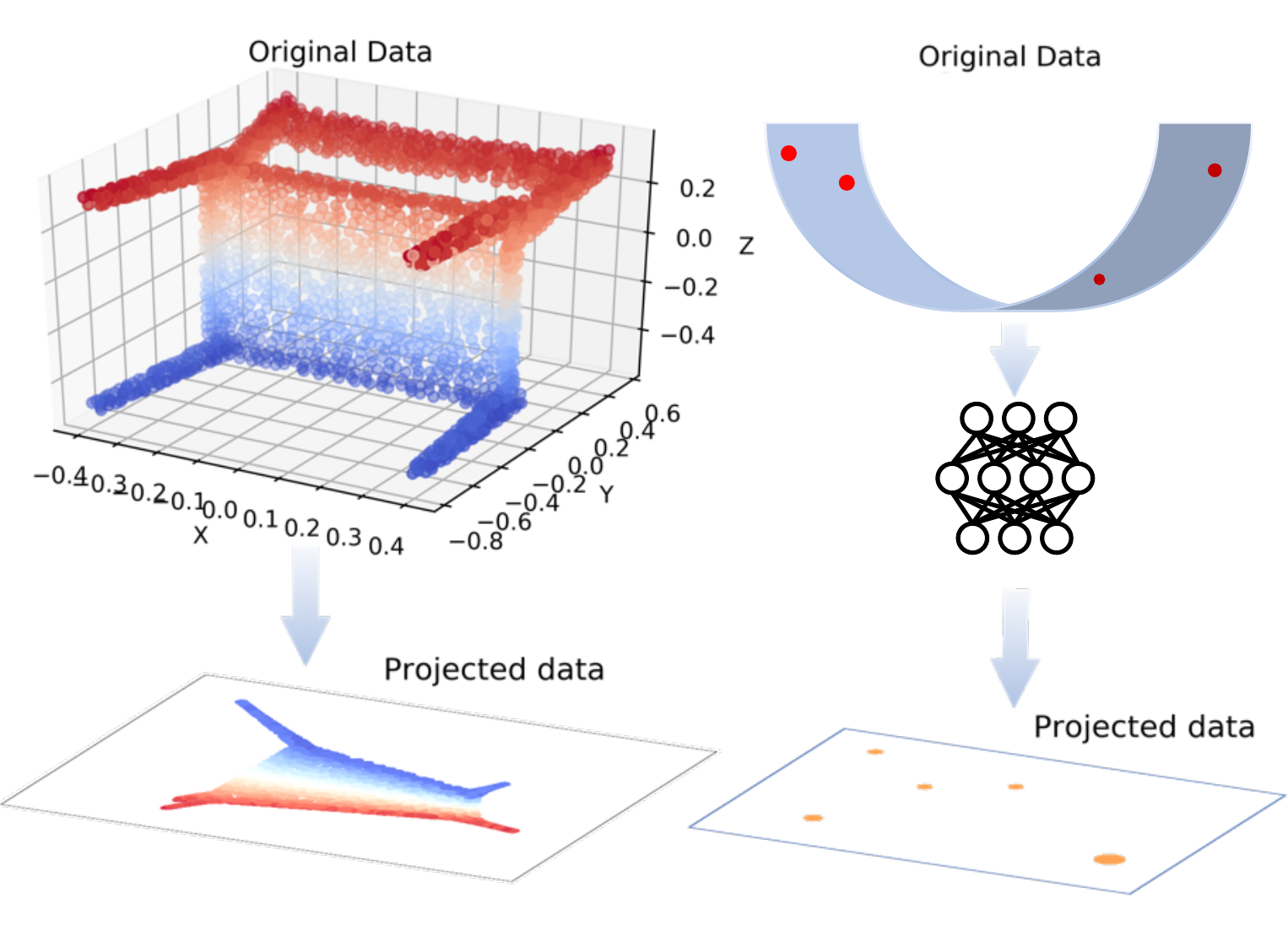}
\end{center}
\caption{Dimension reduction methods for point cloud data. Left: locally linear embedding. Right: neural network based manifold projection.}
\label{fig:LLEandMP}
\end{figure}

%, and it is a generalization of the concepts of surfaces and curves in Euclidean space
Point clouds are mainly used to represent the surface shape of objects. At present, the mainstream methods, e.g. \cite{qi2017pointnet, wang2019dynamic}, directly use Euclidean distance when analyzing the local feature of point clouds, but Euclidean distance cannot accurately reflect the relationship between points due to the curvatures. Additionally, manifold is an extension of curve and surface in the original Euclidean space. Although manifold topological space can be locally treated as Euclidean space, it can be more powerful to evaluate all the elements and their connections. The core idea of manifold learning algorithm is to map data from high dimension to low dimension, which can remove data redundancy while preserving geometric topological relations. Using manifold learning method on point cloud can consider the continuity of the geometric surface of the object, then improve the description for the nature of the geometric shape.

%which can be locally treated as Euclidean space
%Manifold refers to a topological space which can be locally treated as Euclidean space. The core idea of manifold learning algorithm is to map data from high dimension to low dimension, which can remove data redundancy while preserving geometric topological relations. Curves and surfaces belongs to manifold data, 
%and point clouds are mainly used to represent the surface shape of objects. 

%At present, the mainstream methods e.g. \cite{qi2017pointnet, wang2019dynamic} directly use Euclidean space distance when analyzing the local feature of point clouds, but Euclidean distance cannot accurately reflect the relationship between points. 

%Using manifold learning method can overcome this defect, which considering the continuity of the geometric surface of the object, increasing the description of the nature of the geometric shape. 

To introduce manifold learning for point cloud classification, we formulate a novel neural network architecture, named Manifold Learning Module, which consists of two alternative sub-modules, i.e., Locally Linear Embedding (LLE) Module and Manifold Projection (MP) Module. The functions of these two modules are similar. More specifically, the dimensionality of point cloud features is reduced by manifold learning, and then the new generated features are concatenated with the original features to enrich the input features for neural network-based classification. 

In the Locally Linear Embedding Module, we establish the correlation between high-dimensional and low-dimensional spaces by the local symmetries, and then reconstruct the points by neighborhood-preserving mapping. Hence, we can obtain a low-dimensional point set that maintain the continuity of the geometric surface. Additionally, to implement an end-to-end manifold learning method, we present a novel neural network architecture named Manifold Projection Module, which focuses on fitting nonlinear projection mapping from 3D to 2D. Experimental results on the ModelNet40 dataset demonstrate that our methods have better feature representation capability and can lead to better classification results.

% Inspired by \cite{wang2019dynamic}, with the introduction of manifold learning, we propose a novel point cloud analysis method, and achieve better performance than mainstream method in point cloud classification task.

In summary, the main contributions of the proposed method are as follows: 

\begin{itemize}
    \item We propose a novel point cloud classification method with manifold learning and graph neural network, namely PointManifold.
    \item We introduce two manifold learning methods in point cloud classification task, where one is based on locally linear embedding, and the other is a novel manifold projection method based on the designed neural network.
    \item With the feature engineering of manifold learning and feature aggregation of graph neural network, compared with previous PointNet series methods, graph-based methods and other state-of-the-art methods, PointManifold get a competitive performance, and significant improvement from its baseline. Besides, we do some ablation study to explore the relationship between two manifold modules.
\end{itemize}

\section{Related Work}

In this section, we briefly review different categories of deep learning-based point cloud analysis methods.

\textbf{View-based and volumetric methods.} In the task of deep learning-based classification and semantic segmentation of two-dimensional (2D) images, feature extraction and embedding are required. The most direct analysis method is rendering 3D point cloud into 2D images, and then uses conventional 2D image classification neural networks, e.g. \cite{su2015multi, yavartanoo2018spnet, qi2016volumetric}. Moreover, 3D data is a generalization of 2D data, where voxel can be extended from pixel. Therefore, promoting 2D convolutional neural network to 3D convolution is another solution to deal with this task. However, 3D point clouds are sparse and disorderly, and such methods as VoxelNet \cite{maturana2015voxnet} require a large amount of computations. In addition, due to the sparse and uneven density distribution of point clouds, both of view-based and volumetric methods are not sufficiently effective to obtain satisfactory performances. Specifically, large-scale scenes may lead to incredible computation complexity, and the data type transformation from point cloud to voxel will inevitably cause information loss. 

%have a lot of limitations. The analysis performance is poor when encountering large-scale scenes, and some information may be lost in data type transformation.
%Moreover, 3D data is a generalization of 2D data, voxel can be extended from pixel. 
%Moreover, voxel of 3D data can be mapped to be pixel of 2D data. 
%Moreover, the voxels in 3D can be deemed as the pixels in 2D after the dimension space shrinks.

\textbf{Point cloud-based methods (PointNet Series).} To overcome the defects of view-based and volumetric methods, processing each point independently is feasible. The milestone of deep learning-based point cloud analysis is PointNet \cite{qi2017pointnet}, which uses the multi-layer perceptron (MLP) to extract point feature with high dimension, then uses max pooling to obtain the representative feature vector, which solves the disorder of point cloud simultaneously. Compared with the previous methods, PointNet avoids the huge computation of 3D voxel convolution and shows excellent classification and segmentation performances at a higher speed. On the basis of PointNet, \citeauthor{qi2017pointnet++} then propose PointNet++ containing down sampling and up sampling architecture, which enriches the collection of global features and solves the problem of uneven point cloud density. These two methods lay the foundation for the application of deep learning on point cloud analysis in recent years. Additionally, there are derived PointSIFT \cite{jiang2018Pointsift} which focuses on uniform sampling to establish local descriptor, and PointASNL \cite{yan2020pointasnl} which focuses on adaptive sampling.

\textbf{Graph-based methods.} With the development of graph neural network, applying graph to point cloud analysis has become an emerging research direction. There are some representative works \cite{zhang2018PointgCN,wang2019dynamic,li2019deepgcns}. DGCNN \cite{wang2019dynamic} uses graph to express point features and utilizes convolution operation. This work makes a summary for the graph-based point cloud analysis method and expresses the frameworks in formula. 

\textbf{Geometry-based methods.} Geometry-based point cloud analysis network is also developed these years. TangentConv \cite{tatar2018tangent} projects point cloud into tangent planes, then 2D convolution is adopted. FPConv \cite{Lin2020FPConv} learns a local weight matrix to flatten point cloud to 2D grid. MoNet \cite{monti2017geometric} gives a unified framework for generalizing traditional convolutional neural network to non-Euclidean geometric data in spatial domain. ShapeContextNet \cite{Liu2018Attentional} applies the shape context description in traditional computer vision field to point cloud representation. SPLATNet \cite{su2018splatnet} and SO-Net \cite{li2018so} also focus on the representation of point cloud. Meanwhile, \cite{Hermosilla2018Monte, Thomas2019KPConv, Li2018PointCNN, Shen2018mining} focus on migrating convolution operation to point cloud with geometric consideration.

\section{The Proposed Method}
\begin{figure*}
\begin{center}
\includegraphics[width=0.98\linewidth]{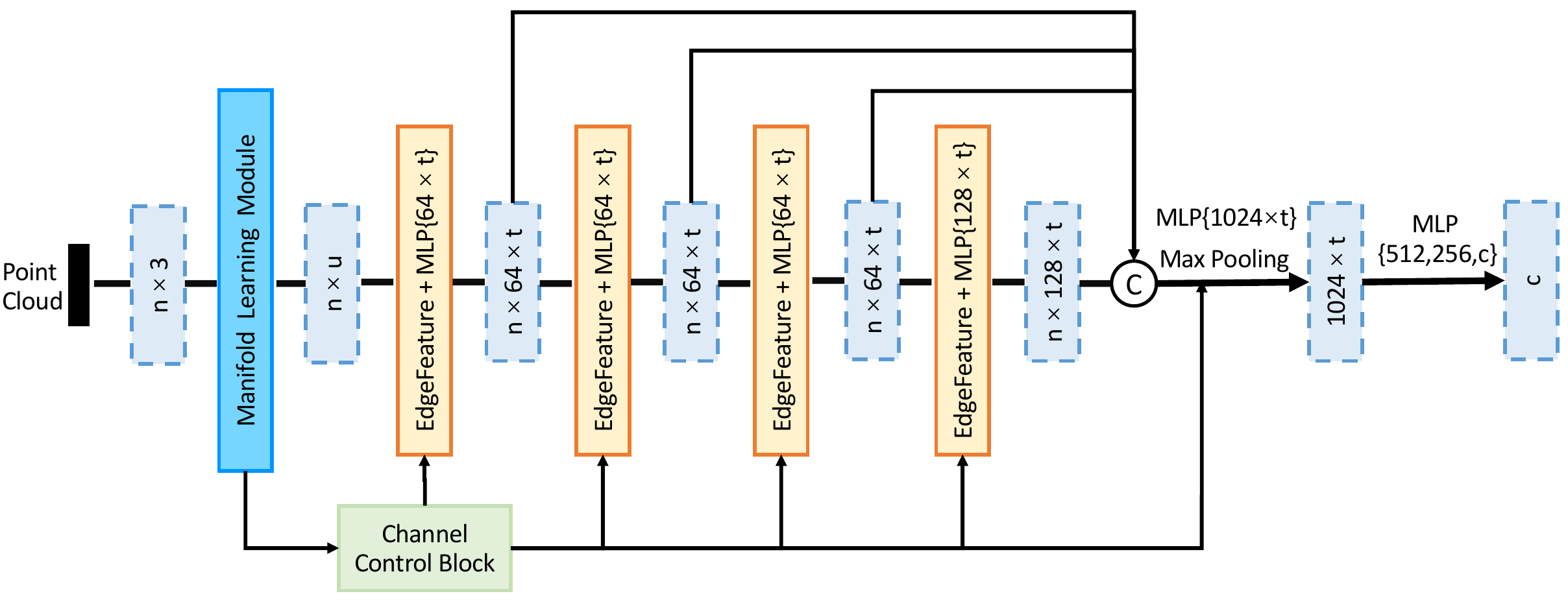}
\end{center}
   \caption{The main architecture of the PointManifold. Manifold Learning Module is consisted of two sub-module: Locally Linear Embedding Module and Manifold Projection Module. MLP is the abbreviation of multi-layer perceptron, and $t$ is the parameter come from Channel Control Block which influence the channel size of each layer. LeakyReLU and BatchNorm are used in each layer, and Dropout layer is set at the last two MLP layers.}
\label{fig:main_architecture}
\end{figure*}

In this section, we introduce PointManifold for point cloud classification. Inspired by DGCNN \cite{wang2019dynamic}, PointManifold uses graph neural network to embed the point features, and then uses convolution operation to extract features for classification task. Meanwhile, to get the geometric nature of point cloud, we apply manifold learning to enrich the dimension of point features. The main architecture of our method is shown in Fig. \ref{fig:main_architecture}. 

\subsection{Locally Linear Embedding Module}

\begin{figure}[h]
\begin{center}
\includegraphics[width=0.85\linewidth]{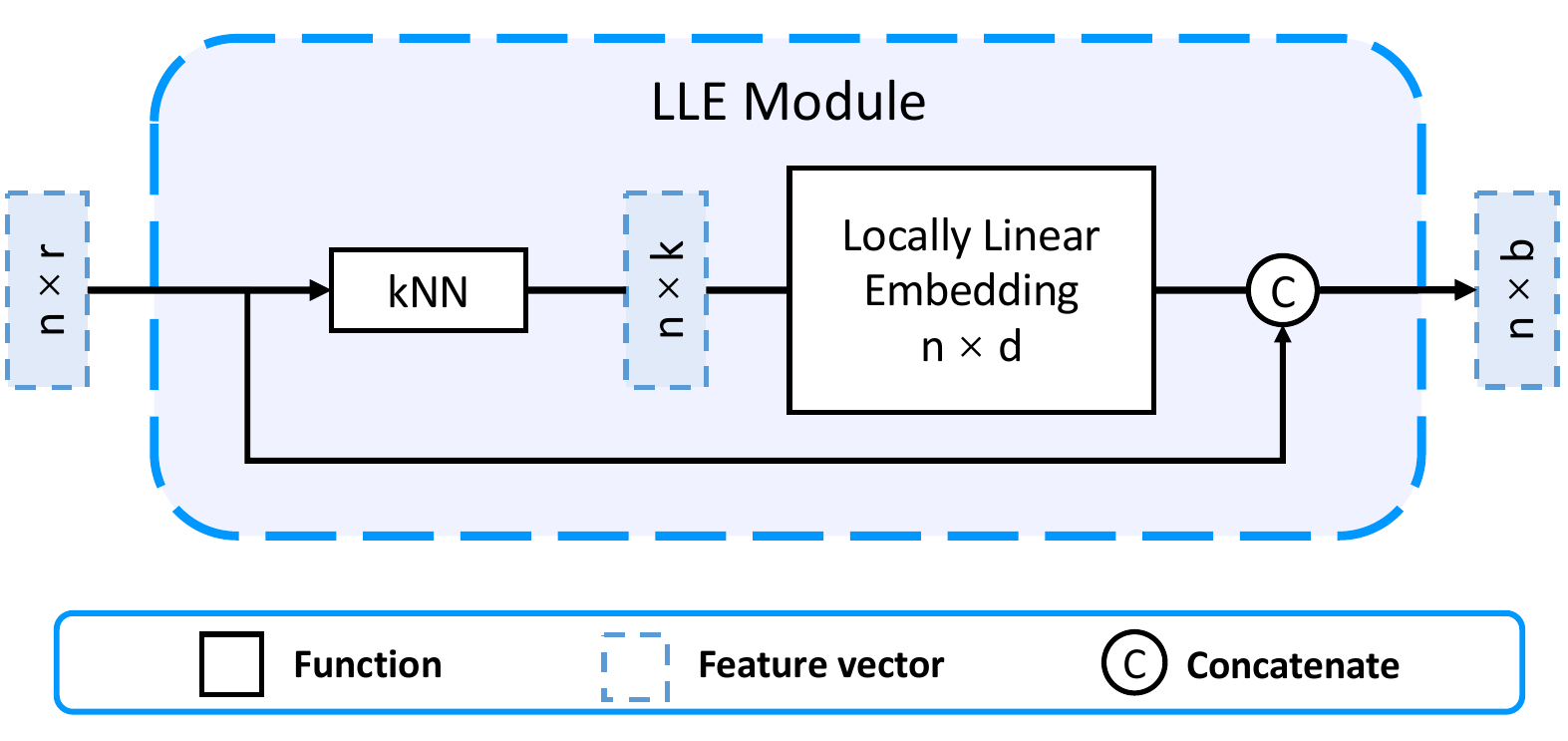}
\end{center}
\caption{The structure of Locally Linear Embedding (LLE) module.}
\label{fig:lle_module}
\end{figure}

%The expression of many data is redundant, some data point may contain thousands of features, while they can be represented by several basic parameters.

In fact, a lot of redundancies still exist in many data representation methods, and each data may even have thousands of features, but they can acutally be well represented by limited parameters. As an efficient method for obtaining these parameters, manifold learning algorithm can give the low dimensional representation for data. There are many methods to solve this problem, such as Isomap, locally linear embedding (LLE) , laplacian eigenmaps, etc. Isomap is a generalization of principal content analysis (PCA) algorithm on manifolds, and it is necessary to calculate the distance between all point pairs. Hence, Isomap requires a large amount of calculation. Since LLE only concerns the equality of distances in local range, it will have less computation overhead. Moreover, LLE can also benefit the local to global neural network-based analysis for point clouds. For the sake of low computation complexity and effectiveness of analysis, we choose locally linear embedding algorithm for point cloud dimension reduction.

\textbf{Notation.} Let $P$ be a point set, $r$ and $d$ are the dimensions of the original space and the new space, respectively, thus $P=\{p_1,...,p_n\}, p_i \in \mathbb{R}^r$. The aim of locally linear embedding is to get a new points representation in lower dimension space, denoted as $\hat P \in \mathbb{R}^{d \times n}, d < r$. 

\textbf{Calculate distance matrix.} The locally linear embedding algorithm requires that the relative distance of the points in the local range should be as unchanged as possible. In other words, one point can be reconstructed by using the features of its adjacent points, which is in fact consistent with the nearest neighbor relationship of point cloud analysis. Let $p_i$ be a central point in local range, $\mathbb{N}_i$ is the $n$ range neighbor of $p_i$, and weighted matrix $W = [w_1,...,w_n], w_i \in \mathbb{R}^n$ is the distance matrix of point pairs. The central point can be represented as:
\begin{equation}
	p_i = \sum_{j \in \mathbb{N}_i} W_{ij}p_j,\ s.t.\sum_j W_{ij} = 1,\forall i \in \{1,...,n\},
\end{equation}

\noindent where $W_{ij}$ denoted the distance weight between $p_i$ and $p_j$, and the constraint is for normalization. $W$ can be obtained by solving:
\begin{equation}
	\mathop{min} \limits_{W} \sum_i || p _i  - \sum_{j \in \mathbb{N}_i} W_{ij} p_j||_2^2.
\end{equation}

\textbf{Get new point representations.} In order to obtain the expression of points in the new feature space $\hat P$, the optimized objective function of locally linear embedding algorithm is:
\begin{equation}
	\mathop{min} \limits_{\hat P} \sum_i || \hat p _i  - \sum_{j \in \mathbb{N}_i} W_{ij} \hat p_j||_2^2,
\end{equation}

\noindent where $\hat p_i$ and $\hat p_j$ are the points of $\hat P$. 

The detailed steps of locally linear embedding algorithm are shown in Alg. \ref{algo:lle}. In the overall network, point cloud data is firstly feed to the devised LLE Module, the structure of which is shown in Fig. \ref{fig:lle_module}. Using LLE to reduce the standardized coordinates to two-dimensional space, the 2D features are then concatenated with the originals. Let $\mathcal{L}$ represent locally linear embedding, $F\in \mathbb{R}^{b \times n}$ be the input feature vector of point set $P$, where $b$ is the feature dimension, and $b = r + d = 5$. We can get the equation of $F$ as:

\begin{equation}
    F_{LLE} = \{\mathbf{x}, \mathbf{y}, \mathbf{z}, \mathcal{L}_{x^{'}}(P),\mathcal{L}_{y^{'}}(P))\},
\end{equation}

\noindent where $x^{'}, y^{'}$ represent the new coordinates of points, and $\mathcal{L} \in \mathbb{R}^{2 \times n}$.

\begin{algorithm}[h]
	\caption{Locally Linear Embedding}
	\label{algo:lle}
	\KwIn{(points set $P$, neighborhood range $K$, new dimension $D$)}
	\KwOut{new representation $Y$ of points set in new space}
	
	\tcc{Using kNN to get the neighborhood}
	\For{Each point $p_i$ in $P$} {
		$N_i = knn(p_i, K)$
	}

	\For{Each point $p_i$ in $P$} {
		$P_i = repeat(p_i, K)$\;
		$S_i = (P_i-N_i)^T(P_i-N_i)$\;
		$w_i = solve\_lagrange\_multipliers(w_i^TS_iw_i + \lambda(w_i^T1_k - 1))$\;
	}

	\tcc{Solve for a low-dimensional mapping}
	$W' = new\_matrix(N, N)$\;
	\For{Each point $p_i$ in $P$} {
	\For{$j\ in\ range(N)$} {
		\If{$j \in N_i$}{
			$W'_{ji} = w_{ji}$\;
		}
		\Else{
			$W'_{ji} = 0$\;
		}
	}
	}

	\tcc{The minimum distance loss is obtained by using the Lagrange multiplier method, in this case, $Y$is a matrix composed of the eigenvectors of $M$}
	$Y = new\_matrix(D, N)$\;
	
	\For{$i\ in\ range(N)$}{
		$M_i = (I_i - W_i)(I_i - W_i)^T$\;
		$eigen\_vec, eigen\_val = eigens(M_i, D)$\;
		$Y_i = eigen\_vec(2:D+1)$\;
	}

	\Return{$Y$}
\end{algorithm}

\subsection{Manifold Projection Module}
The core idea of manifold learning algorithm is to find a mapping from high-dimensional space to low-dimensional space. Different from the traditional manifold learning algorithms, which use the distance measurement in different spaces to reduce dimension, we design a novel neural network architecture for this task.

\begin{figure}[ht]
	\begin{center}
		\includegraphics[width=0.8\linewidth]{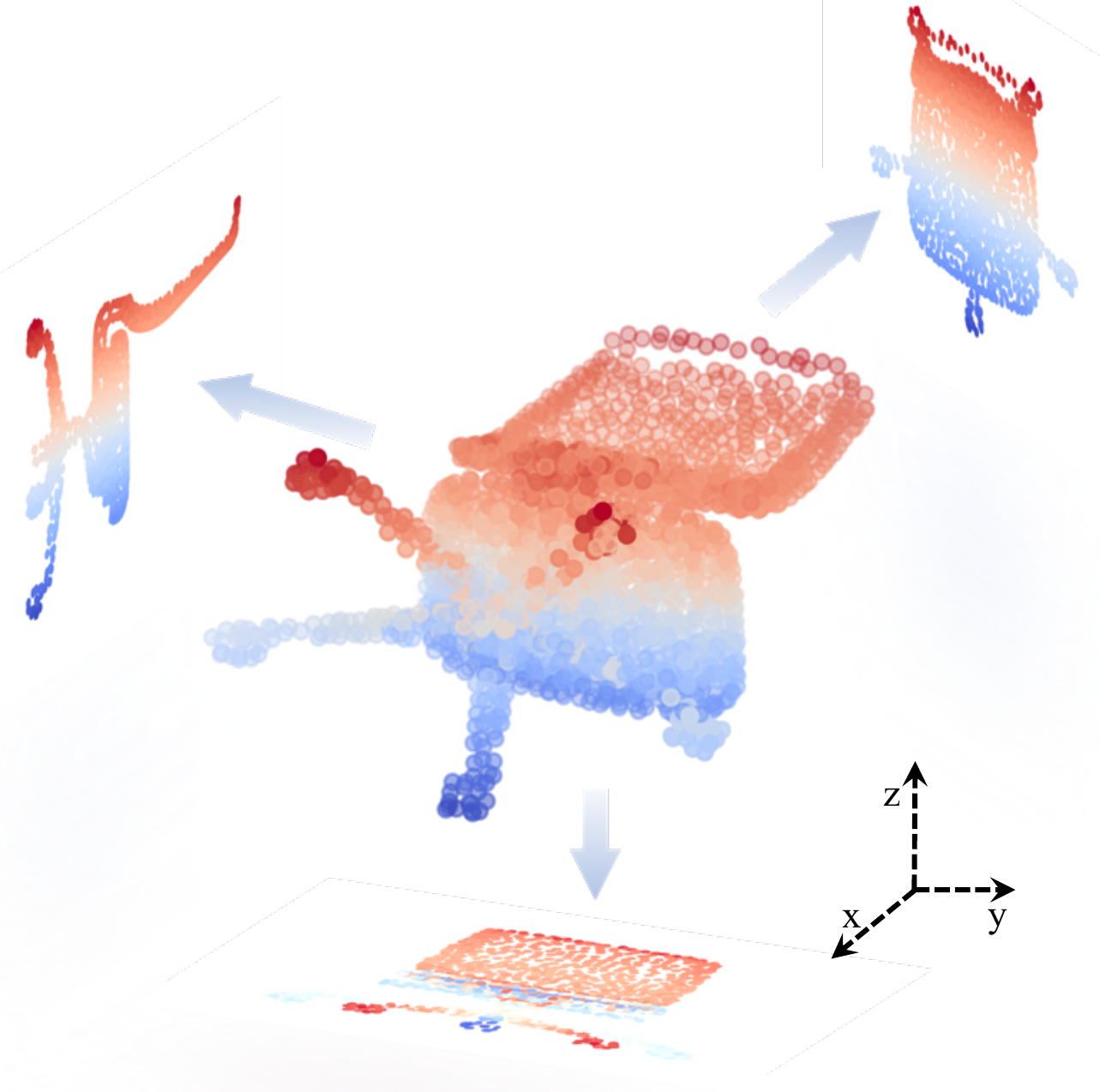}
	\end{center}
	\caption{Project point cloud into three plane $x=0$, $y=0$, 
		$z=0$ by MP module.}
	\label{fig:projection}
\end{figure}

\textbf{Projection in Euclidean space.} Let $Ax + By + Cz + D = 0$ be a plane in 3-dimensional Euclidean space, $p_i(x_i,y_i,z_i)$ is a point, $p_i^{'}(x_i^{'},y_i^{'},z_i^{'})$ is the projection of point $p_i$ on the plane. According to the vertical relation, the parametric equation of the vertical line is:
\begin{equation}
\label{eq:projection}
	\cases{x_i^{'}=x_i-At  \cr y_i^{'}=y_i-Bt  \cr z_i^{'}=z_i-Ct}.
\end{equation}

Since $p_i^{'}$ is on the plane, $t$ can be solved by the plane equation:
\begin{equation}
\label{eq:t}
    t= \frac{Ax_i+By_i+Cz_i+D}{A^2+B^2+C^2}.
\end{equation}

\textbf{Nonlinear projection by neural network.}
We can get the linear projection function $S$ by Eq.\ref{eq:projection} and Eq.\ref{eq:t}. As manifold learning is a nonlinear dimensionality reduction method, we set a nonlinear function $\mathcal{S}$ for this process:
\begin{equation}
\label{eq:mpf}
    \mathcal{S}_{\beta}(x,y,z) = \mathcal{X}_{\beta}(x,y,z)S,
\end{equation}

\noindent where $\beta$ is the projection plane, and $\mathcal{X}$ is a nonlinear function defined as follows:

\begin{equation}
    \mathcal{X}(\cdot) = \mathcal{F}(Q_{\Theta}(\cdot)),
\end{equation}

\noindent where $\mathcal{F}$ is an activation function, and $Q_{\Theta}$ is a function to fit the mapping. Here, we implement it by multi-layer perceptron. Then, we can get the full definition of manifold projection $\mathcal{S}$:
\begin{equation}
     \mathcal{S}_{\beta}(x,y,z) = \mathcal{F}(Q_{\Theta}(x, y, z, \beta))S_{\beta}(x, y, z).
\end{equation}

\begin{figure}[h]
\begin{center}
\includegraphics[width=0.85\linewidth]{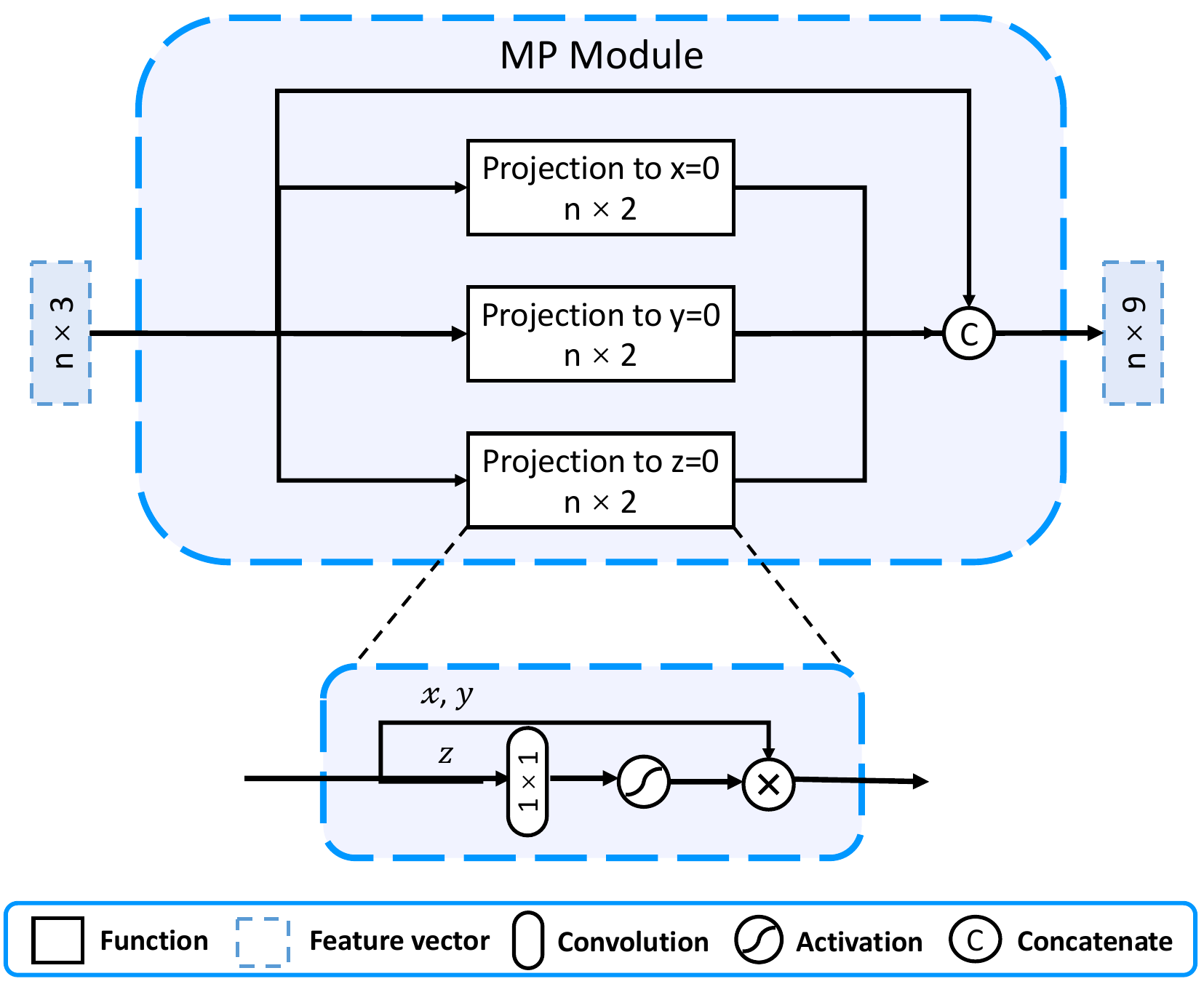}
\end{center}
\caption{The structure of Manifold Projection (MP) module.}
\label{fig:mp_module}
\end{figure}

Finally, since all data of ModelNet40 are standardized, we take three projection planes $x=0, y=0, z=0$ to get a multi-view of the point cloud, as shown in Fig.\ref{fig:projection}. Similar with the LLE Module, we concatenate the new manifold features with the original features. Then, we can get the input feature vector $F$ as:
\begin{equation}
    F_{MP} = \{\mathbf{x}, \mathbf{y}, \mathbf{z}, \mathcal{S}_{x=0}(P),\mathcal{S}_{y=0}(P), \mathcal{S}_{z=0}(P)\},
\end{equation}

\noindent where $\mathcal{S} \in \mathbb{R}^{6 \times n}$. here $d = 2$, thus the dimension of new feature $b = r + 3 \times d = 9$. The architecture of manifold projection module is shown in Fig.\ref{fig:mp_module}.
\subsection{EdgeConv Module}
\begin{figure}[h]
	\begin{center}
		\includegraphics[width=0.85\linewidth]{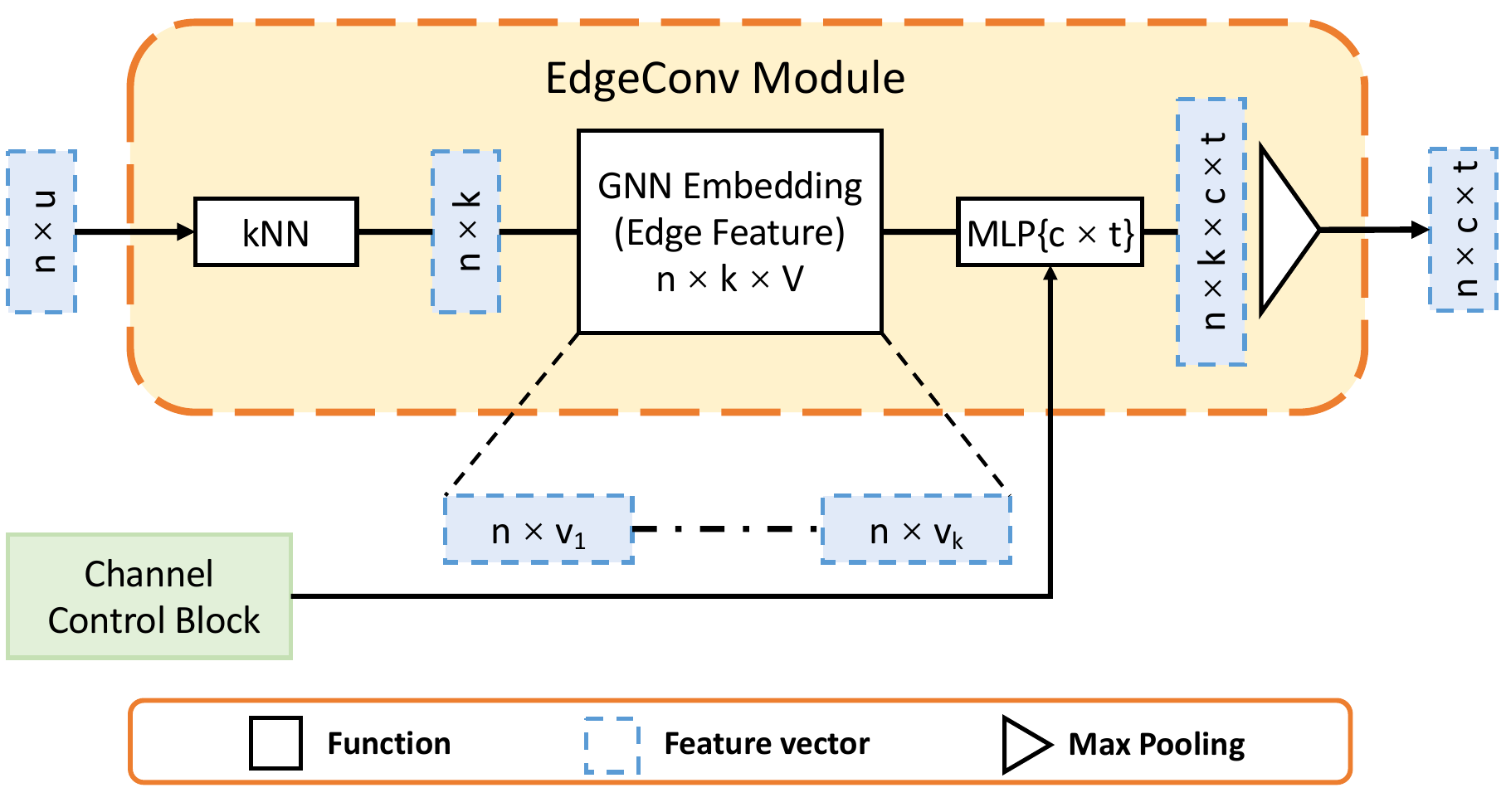}
	\end{center}
	\caption{The structure of EdgeConv module. The kNN algorithm get $k$ nearest neighbors for each point, the graph neural network establish edge feature $v$ between each point and its neighbors.}
	\label{fig:edge_conv}
\end{figure}

The EdgeConv Module is mainly based on DGCNN \cite{wang2019dynamic}, and the structure is shown in Fig. \ref{fig:edge_conv}. Let $f_i$ be the feature vector of central point $i$, $\mathbb{N}_i$ be the neighbor of $i$, which is selected by k nearest neighbor (kNN). We denote $e_{ij}$ as the edge feature of $i$ and its neighbor $j$, and the definition is:
\begin{equation}
	e_{ij} = h_\Theta(f_i, f_j), j \in \mathbb{N}_i,
\end{equation}

\noindent where $h_\Theta(\cdot, \cdot)$ is the edgeconv function. We implement it by multi-layer perceptron. Then, we can get the new feature vector $f_i^{'}$:

\begin{equation}
	f_i^{'} = g(e_{ij}) = \mathop{g} \limits_{j\in\mathbb{N}_i}(h_\Theta(f_i, f_j)),
\end{equation}

\noindent where $g(\cdot)$ is a symmetric function, and here we use maximum. Finally, we achieve the full definition of edgeconv:
\begin{equation}
	f_i^{'} = \mathop{max} \limits_{j\in\mathbb{N}_i}(LeakyReLU(\theta_m \cdot f_i + \phi_m \cdot (f_j - f_i))),
\end{equation}

\noindent where $\Theta = (\theta_1,...,\theta_M,\phi_1,...,\phi_M)$ is the parameters to be learnt, and $M$ is the number of convolution kernels.

Since the dimension of input feature vector is higher than that in \citet{wang2019dynamic}, we add a dynamic channel control block on the backbone network, which can adjust the channel size of each layer to fit different dimensions of input data. In Experiments section, we will analyse the influence of channel size on classification performance.
% \FloatBarrier
\section{Experiments}
In this section, to demonstrate the effectiveness of PointManifold, we design a set of experiments on ModelNet40 \cite{wu20153d}, and compare PointManifold with a series of state-of-the-art methods. Additionally, we do several ablation experiments to deeply explore the operation mechanism of Manifold Learning Module and Channel Control Block.
 % We also test the semantic segmentation performance of our methods on Stanford Large-Scale 3D Indoor Space (S3DIS) \cite{armeni_cvpr16} dataset.

\subsection{Classification on ModelNet40}

\textbf{Data.} We select ModelNet40 for the classification task, which contains 12308 models and a total of 40 categories. All the models are sampled from the computer-aided design (CAD) models. Each model provides 2048 sampling points pre-processed with standardization, and each point contains 3D position information. In training process, 9840 models are used as the training set, and the remaining 2468 models are used as the test set. In order to speed up the training process and avoid repeated processing of the same data between epochs, we use locally linear embedding algorithm to pre-process the data. This work can reduce the standardized coordinates to two-dimensional space, and then the new features can be concatenated to the original features.

\textbf{Environment.} The code implementation of the proposed method is based on Pytorch framework (version 1.1.0) and Python (version 3.6). The experimental computing platform includes one NVIDIA RTX 2070 and four NVIDIA Tesla V100. The operating system is Ubuntu (version 16.04), CUDA (version 10.1) and cuDNN (version 7.4). 

\textbf{Hyper-parameters.} We use SGD as optimizer with a momentum of 0.9, and the learning rate is 0.1 with a cosine annealing scheduler. To enhance the fitting ability of the proposed model, we add dropout layer with 0.5 dropout ratio. For the classification activation function, we select softmax. For LLE Module, we set training epochs as 250, batch size as 32, channel time $t$ as 1, and the neighbor range $k$ of edgeconv is 20, while the $k$ of LLE is 12. Specifically, both of the range parameters will be doubled in 2048 points experiments. For MP Module, our model is trained with 300 epochs, and the channel time $t$ are set as 2 and 4 for 1024 and 2048 points experiments, respectively.

\textbf{Result.} Following \citet{wang2019dynamic}, we report the mean class accuracy (mA) and overall accuracy (oA) on ModelNet40. Results are shown in Table.\ref{tab:modelnet40}, as we can see the proposed methods achieve a competitive result. PointManifold is better than the mainstream state-of-the-art point cloud analysis methods in classification task, and have a significant improvement compared with the DGCNN baseline. In addition, we test the class-level result for point cloud classification and give more explorations about the relationship between shape and model performance, which is provided in the Supplementary Material. For fair comparison, the results of DGCNN is tested by ourselves, with the same hyper-parameters settings as its open source code \cite{wang2019dynamic}, and the same environment as our methods. Besides, because of the limitation of our experiment platform, we cannot train MP Module (2048 points) with the batch size of 32, and we set it as 24 instead. We believe that the classification performance could be better if the batch size is larger.

\begin{table}[h]
	\centering
	\begin{tabular}{lll}
	\toprule
	Method & \tabincell{l}{Mean Class \\ Accuracy} & \tabincell{l}{Overall \\ Accuracy}\\
	\midrule
	3D ShapeNet [\citeyear{wu20153d}] & 77.3 & 84.7\\
	VoxelNet [\citeyear{maturana2015voxnet}] & 83.0 & 85.9\\
	MVCNN [\citeyear{su2015multi}] & 79.5 & 90.1\\
	PointNet [\citeyear{qi2017pointnet}] & 86.0 & 89.2\\
	PointNet++ [\citeyear{qi2017pointnet++}] & - & 90.7\\
	PointCNN [\citeyear{Li2018PointCNN}] & 88.1 & 92.2\\
	SPNet [\citeyear{yavartanoo2018spnet}] & 85.2 & 92.6\\
	PointConv[\citeyear{Wu_2019_CVPR}] & - & 92.5\\
	DGCNN (1024 points)[\citeyear{wang2019dynamic}] & 89.3 & 92.7\\
	DGCNN (2048 points) & 90.1 & 92.8\\
	KPConv [\citeyear{Thomas2019KPConv}] & - & 92.9\\
	FPConv [\citeyear{Lin2020FPConv}] & - & 92.5\\
	Point2Node [\citeyear{han2020point2node}] & - & 93.0\\
	\midrule
	Ours (LLE, 1024 points) & \textbf{90.0} & \textbf{93.0}\\
	Ours (LLE, 2048 points) & \textbf{90.2} & \textbf{93.2}\\
	\midrule
	Ours (MP, 1024 points) & \textbf{90.1} & \textbf{93.0}\\
	Ours (MP, 2048 points) & \textbf{90.1} & \textbf{93.1}\\
	\bottomrule
	\end{tabular}
	\caption{Classification result on ModelNet40. All the methods without special annotations are sampled at 1024 points.}
	\label{tab:modelnet40}
\end{table}

\subsection{Manifold Learning Effectiveness} 
PCA is a common dimension reduction method, which computes principal components and uses them to perform the changes on the basis of the data. In order to verify the effectiveness of manifold learning in geometric feature extraction, we conduct a controlled experiment between PCA and our methods, and the results are shown in Table. \ref{tab:pca}. The PCA algorithm makes no improvement compared with the baseline, while LLE and MP module can take into account the geometric properties when reducing dimensions. Hence, they can perform better than the DGCNN baseline. 

\begin{table}[h]
	\centering
	\begin{tabular}{lll}
	\toprule
    Method & \tabincell{l}{Mean Class \\ Accuracy} & \tabincell{l}{Overall \\ Accuracy}\\
	\midrule
	DGCNN [\citeyear{wang2019dynamic}] & 89.3 & 92.7\\
	\midrule
	Ours (with PCA) & 89.2 & 92.5 \\
	Ours (with LLE) & \textbf{90.0} & \textbf{93.0}\\
	Ours (with MP on 1 plane) & \textbf{89.9} & \textbf{92.8}\\
	Ours (with MP on 3 planes) & \textbf{90.1} & \textbf{93.0}\\
	\bottomrule
	\end{tabular}
	\caption{ModelNet40 classification result with different methods of dimension reduction (1024 points).}	
	\label{tab:pca}
\end{table}
 
\subsection{Ablation Study} 
 As mentioned above, we add a channel control block, thus we set an ablation experiment to explore the influence of channel control parameter $t$ on the model performance. In addition, we also would like to investigate the relationship between the two manifold learning methods, and therefore we set up several sets of ablation experiments.
 
 From Table.\ref{tab:ablation}, it can be seen that, with the additional input features, an appropriate increase in the channel numbers of the entire model can improve the classification performance. Furthermore, due to the function overlapping, using LLE and MP modules for feature extraction simultaneoulsy cannot fully improve the classification performance, and using either of them can get a similar improvement compared with the baseline. In other words, the ablation study results prove that the proposed neural network architecture can implement the function of manifold learning. 

 \begin{table}[h]
	\centering
	\begin{tabular}{llllll}
	\toprule
    LLE & MP & Planes & \tabincell{l}{t} & \tabincell{l}{Mean Class \\ Accuracy} & \tabincell{l}{Overall \\ Accuracy}\\
	\midrule
	 & & - & 1 & 89.3 & 92.7\\
	\checkmark &  & 1 & 1 & 90.0 & \textbf{93.0}\\
	 & \checkmark & 1 & 1 & 88.9 & 92.5\\
	 &  \checkmark & 3 & 2 & 89.7 & 92.5\\
	 &  \checkmark & 3 & 4 & \textbf{90.1} & \textbf{93.0}\\
	\checkmark & \checkmark & 3 & 4 & 89.5 & 92.7\\
	\bottomrule
	\end{tabular}
	\caption{Ablation study result on ModelNet40 (1024 points).}	
	\label{tab:ablation}
\end{table}

\section{Conclusion}
In this work, we propose a novel point cloud classification method with manifold learning and graph neural network. Using the proposed locally linear embedding algorithm or nonlinear neural network projection, we can get low dimension point features according to the geometric correlation. These new features are concatenated with the original 3D features, and we extracted them by using graph neural network. The experiments demonstrate that our methods with different manifold learning strategies can improve the point cloud classification performances when compared with the DGCNN baseline method. It can be seen that, in the proposed method, the effectiveness of extracted features are guaranteed by considering the surface continuity of point cloud during the 3D-to-2D projection. Therefore, we believe that, besides manifold learning, many other feature analysis approaches will also be very helpful for the feature representation learning in the point cloud data analysis and understanding tasks.  

%projection for generating more efficient features
%these features can reflect more information about the surface continuity
%in the proposed method, the effectiveness of extracted features are guaranteed by considering the surface continuity of point cloud during the 3D-to-2D projection. 

%Moreover, this thinking may be a general improvement for point cloud classification task, from 3D to 2D, or even multi-view perspective, may make a different solution for point cloud semantic analysis.

% \bibliographystyle{aaai21}
%\bibliography{egbib}

\clearpage
\section{Supplementary Material}
\subsection{More Results on Classification Task}
In order to further explore the effectiveness of Manifold Learning Module on classification task, we test the classification performance of 40 categories of ModelNet40 in detail, and we report three metrics for each category, i.e., precision, recall and f1-score, the results are shown in Table.\ref{tab:classification_detail}, from which we can see:
\begin{itemize}
    \item In most categories, compared with the DGCNN\cite{wang2019dynamic} baseline, our methods make improvement or achieve the same result. However, in four categories, i.e., bench, curtain, door and night\_stand, our approaches are not better than the baseline.
    \item On some of the categories in which the physical features are prominent, our methods make significant improvement, such as bookshelf, cup, dresser, radio, sink and xbox.
    \item The classification performance of flower\_pot is extremely poor, in spite of 4\% improvement by MP Module. We visualize some samples to find out the reason.
\end{itemize}

\subsection{Visualization and Analysis}
We attempted to explain the above results by visualization.

\subsubsection{Shape-similar Categories.}
The classification of shape-similar categories is difficult. To explain the bad classification performance of flower\_pot category, we calculate the confusion matrixes of the baseline and our methods, and the results of specific classes are shown in Table.\ref{tab:flower_pot}. 

\begin{table}[h]
	\centering
	\begin{tabular}{lllllll}
	\toprule
    Method & \tabincell{l}{flower\\pot} & \tabincell{l}{book-\\shelf} & bowl & cup & plant & vase\\
	\midrule
	DGCNN & 0.15 & 0.05 & 0.00 & 0.00 & 0.55 & 0.25\\
	\midrule
	LLE & 0.05 & 0.00 & 0.05 & 0.05 & 0.65 & 0.20 \\
	MP & \textbf{0.20} & 0.00 & 0.00 & 0.00 & 0.55 & 0.25\\
	\bottomrule
	\end{tabular}
	\caption{Prediction recall on flower\_pot category (1024 points).}	
	\label{tab:flower_pot}
\end{table}

\begin{figure}[h]
	\begin{center}
		\includegraphics[width=0.7\linewidth]{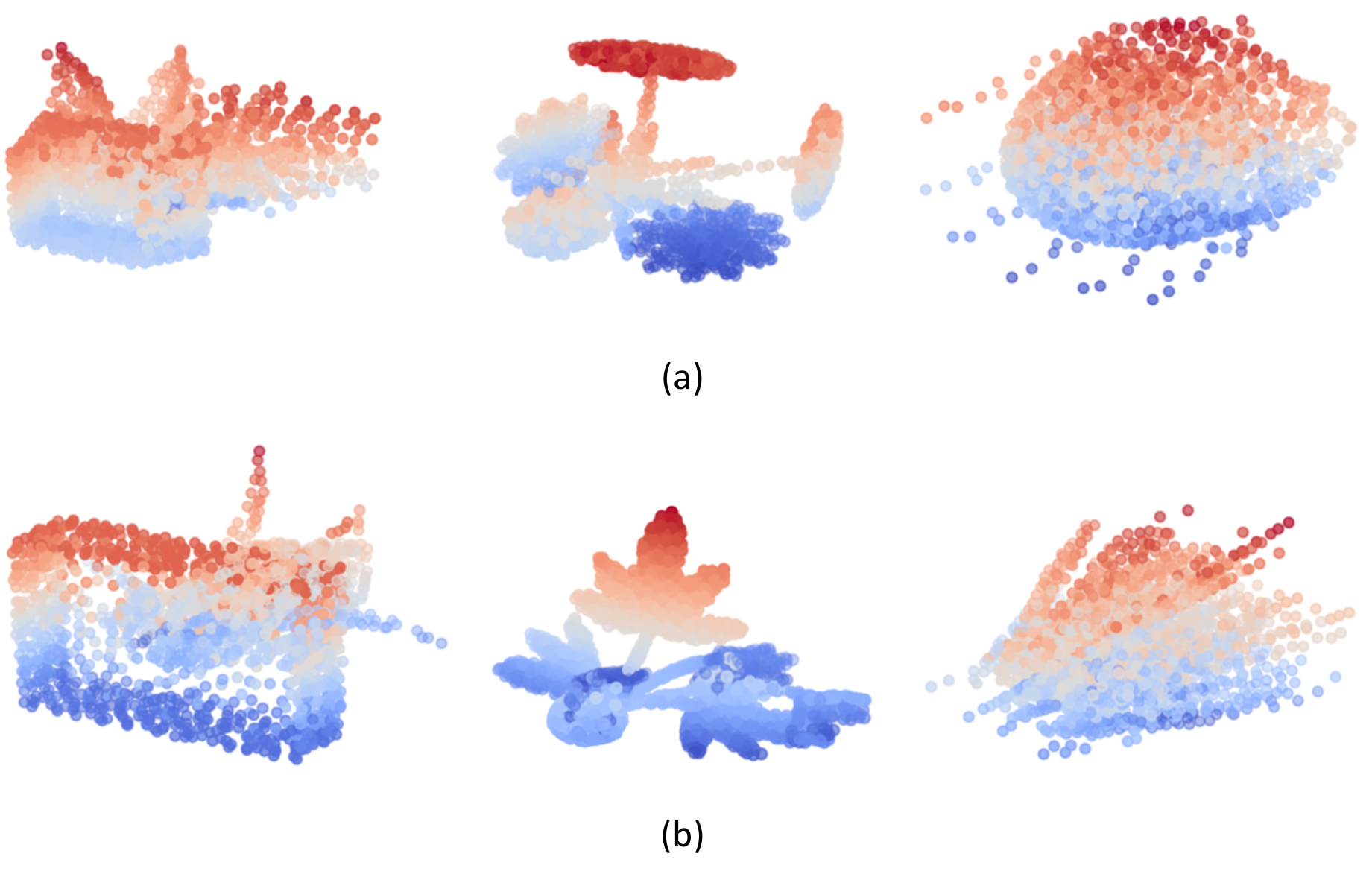}
	\end{center}
	\caption{Samples of ModelNet40: (a) flower\_pot, (b) plant.}
	\label{fig:flower_pot}
\end{figure}

\begin{figure}[h]
	\begin{center}
		\includegraphics[width=0.7\linewidth]{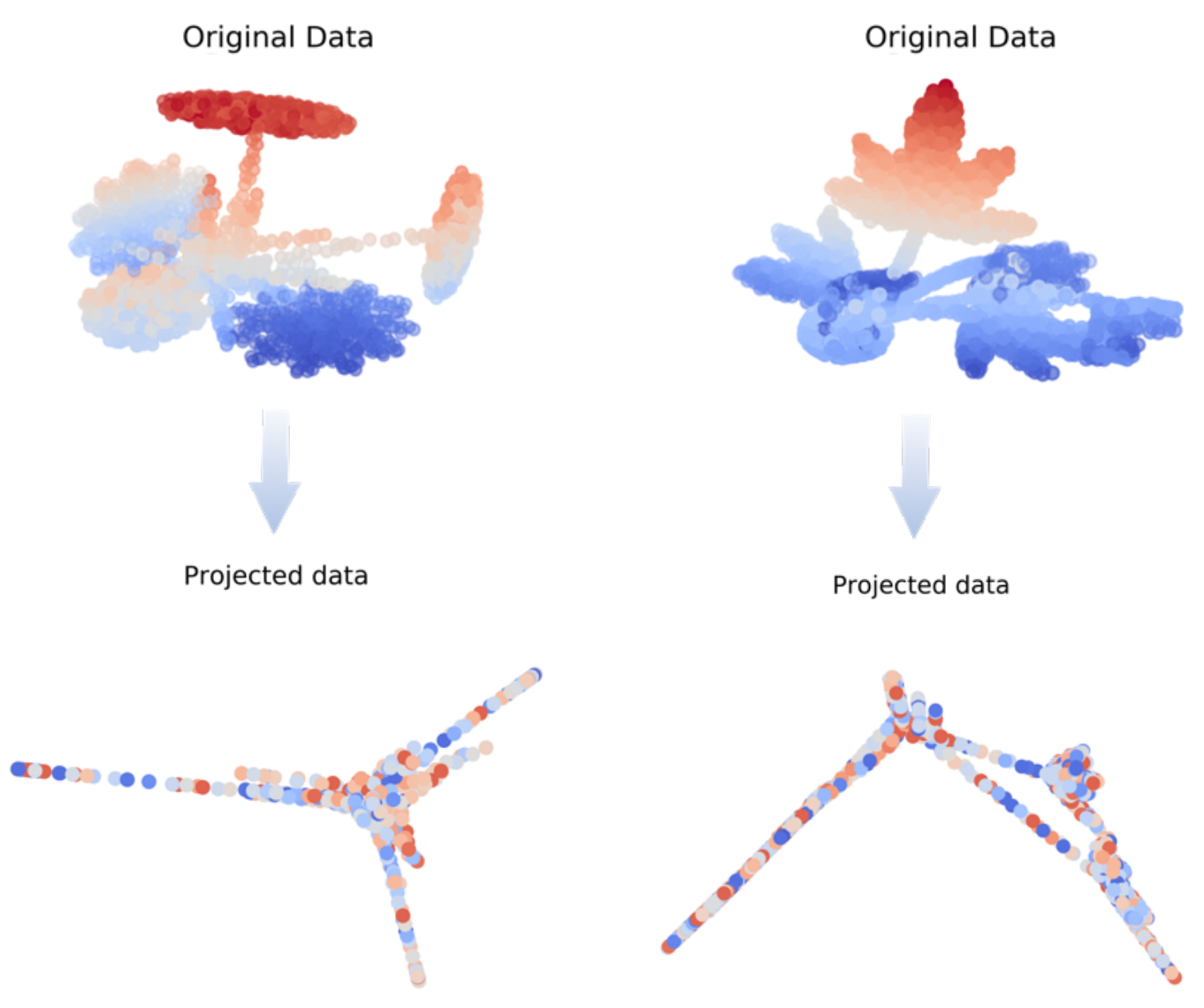}
	\end{center}
	\caption{LLE samples. Left: flower\_pot . Right: plant.}
	\label{fig:flower_pot_lle}
\end{figure}

For all the three methods, more than half of flower\_pot samples are classified as plant. We visualize the two categories, and select some representative samples to show in Fig. \ref{fig:flower_pot}. We find that some of them are nearly the same, since some plants are with flower\_pots, and some flower\_pots are not get rid of the plants. The samples of the two categories are similar, thus the features of them extracted by locally linear embedding in 2D space is also difficult to classify, as shown in Fig.\ref{fig:flower_pot_lle}. 

In summary, LLE Module is not suitable for the shape-similar categories, while MP Module can overcome this defect by multi-view feature concatenation. In addition, the 2D feature are easily confused between the shape-similar categories, especially those whose shapes are regular and concise, thus the classification performance is not improved on bench, curtain, door and night\_stand.

\subsubsection{Feature-prominent Categories.}
\begin{figure}[h]
	\begin{center}
		\includegraphics[width=0.7\linewidth]{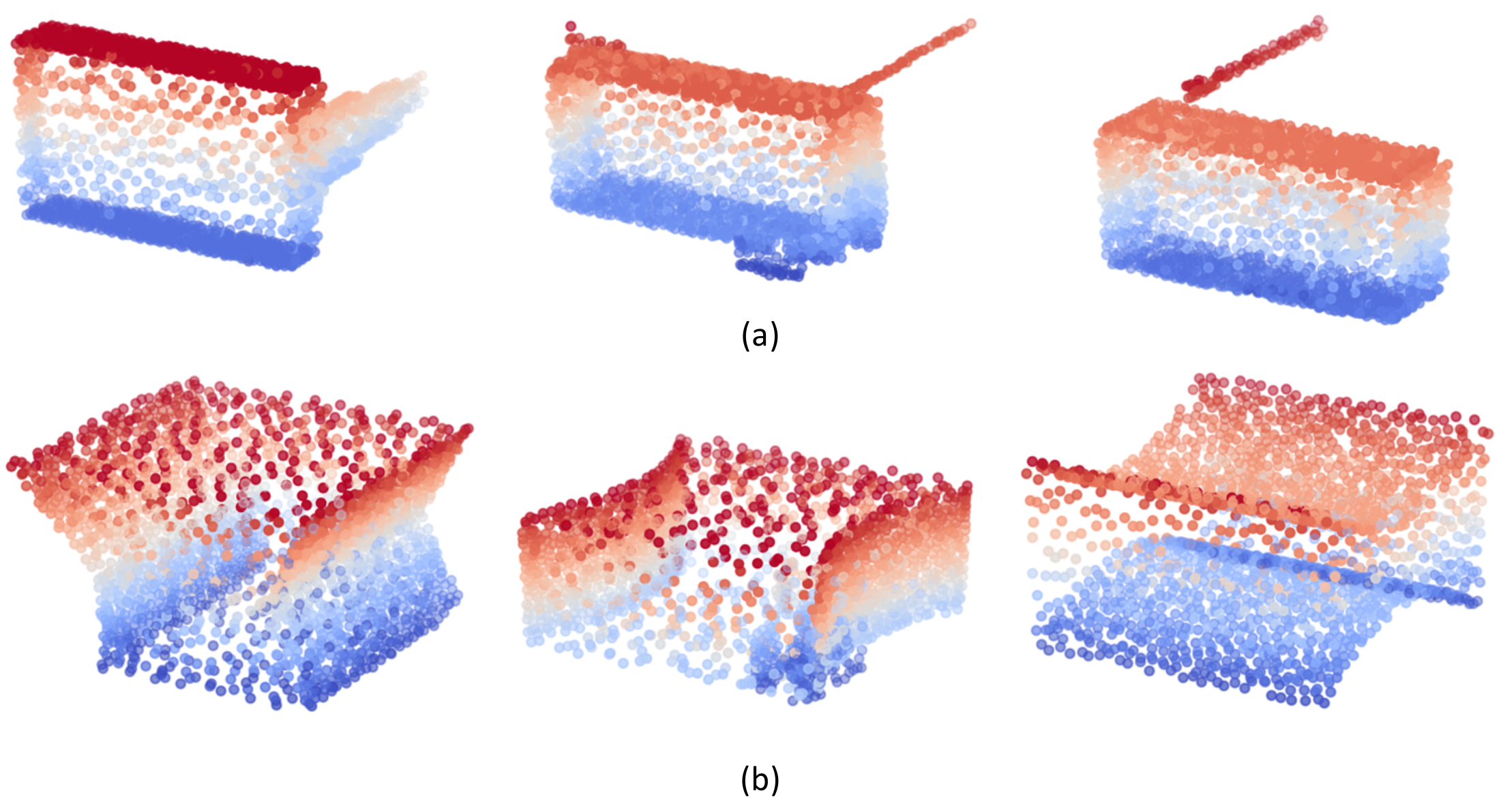}
	\end{center}
	\caption{Samples of ModelNet40: (a) radio. (b) xbox.}
	\label{fig:radio_and_xbox}
\end{figure}

On the contrary, with the Manifold Learning Module, the classification effect of the categories with prominent shape features (e.g. antenna, camber) is improved significantly, which proves our methods can extract more geometric features of 3D models. For example, the radio and xbox make an obvious improvement, and we select some representative samples to show in Fig.\ref{fig:radio_and_xbox}.

\begin{table*}[h]
\centering
\begin{tabular}{l|l|cccccccc}
\toprule
Method & Metric &
airplane &
bathtub &
bed &
bench &
bookshelf &
bottle &
bowl &
car
\\
\midrule
\multirow{3}{*}{DGCNN[\citeyear{wang2019dynamic}]} 
& precision & 1.00 & 0.98 & 0.98 & 0.84 & 0.90 & 0.97 & 0.86 & 0.98\\ 
& recall & 1.00 & 0.96 & 0.99 & 0.80 & 0.99 & 0.97 & 0.90 & 1.00\\
& f1-score & \textbf{1.00} & \textbf{0.97} & 0.99 & \textbf{0.82} & 0.94 & 0.97 & \textbf{0.88} & 0.99\\
\cline{1-2} 
\multirow{3}{*}{Ours (LLE)} 
& precision & 1.00 & 1.00 & 0.99 & 0.88 & 0.94 & 0.98 & 0.77 & 0.99\\
& recall & 1.00 & 0.94 & 0.99 & 0.75 & 1.00 & 0.98 & 1.00 & 1.00\\
& f1-score & \textbf{1.00} & \textbf{0.97} & \textbf{0.99} & 0.81 & \textbf{0.97} & \textbf{0.98} & 0.87 & \textbf{1.00}\\
\cline{1-2} 
\multirow{3}{*}{Ours (MP)} 
& precision & 1.00 & 0.98 & 0.99 & 0.71 & 0.93 & 0.98 & 0.86 & 0.99\\
& recall & 1.00 & 0.96 & 0.99 & 0.75 & 0.99 & 0.97 & 0.90 & 1.00\\
& f1-score & \textbf{1.00} & \textbf{0.97} & \textbf{0.99} & 0.73 & 0.96 & 0.97 & \textbf{0.88} & \textbf{1.00}\\
\midrule
\midrule
Method & Metric &
chair &
cone &
cup &
curtain &
desk &
door &
dresser &
flower\_pot\\
\midrule
\multirow{3}{*}{DGCNN[\citeyear{wang2019dynamic}]} 
& precision & 0.98 & 1.00 & 0.65 & 0.95 & 0.86 & 0.90 & 0.81 & 0.14\\ 
& recall & 0.98 & 0.95 & 0.65 & 0.95 & 0.91 & 0.95 & 0.84 & 0.15\\
& f1-score & 0.98 & \textbf{0.97} & 0.65 & \textbf{0.95} & \textbf{0.88} & \textbf{0.93} & 0.82 & 0.15\\
\cline{1-2} 
\multirow{3}{*}{Ours (LLE)} 
& precision & 0.97 & 1.00 & 0.70 & 0.86 & 0.82 & 0.95 & 0.79 & 0.07\\
& recall & 0.98 & 0.95 & 0.70 & 0.95 & 0.90 & 0.90 & 0.90 & 0.05\\
& f1-score & 0.98 & \textbf{0.97} & \textbf{0.70} & 0.90 & 0.86 & 0.92 & 0.84 & 0.06\\
\cline{1-2} 
\multirow{3}{*}{Ours (MP)} 
& precision & 0.99 & 1.00 & 0.65 & 0.86 & 0.84 & 0.90 & 0.86 & 0.17\\
& recall & 0.98 & 0.95 & 0.65 & 0.95 & 0.93 & 0.90 & 0.90 & 0.20\\
& f1-score & \textbf{0.98} & \textbf{0.97} & 0.65 & 0.90 & \textbf{0.88} & 0.90 & \textbf{0.88} & \textbf{0.19}\\
\midrule
\midrule
Method & Metric &
glass\_box &
guitar &
keyboard &
lamp &
laptop &
mantel &
monitor &
night\_stand\\
\midrule
\multirow{3}{*}{DGCNN[\citeyear{wang2019dynamic}]} 
& precision & 0.97 & 0.99 & 0.95 & 1.00 & 0.95 & 0.99 & 0.97 & 0.81\\ 
& recall & 0.97 & 1.00 & 0.95 & 0.85 & 1.00 & 0.96 & 1.00 & 0.80\\
& f1-score & 0.97 & \textbf{1.00} & \textbf{0.95} & \textbf{0.92} & 0.98 & 0.97 & 0.99 & \textbf{0.81}\\
\cline{1-2} 
\multirow{3}{*}{Ours (LLE)} 
& precision & 0.98 & 0.98 & 0.95 & 0.94 & 0.95 & 0.99 & 0.97 & 0.86\\
& recall & 0.96 & 1.00 & 0.95 & 0.85 & 1.00 & 0.96 & 1.00 & 0.72\\
& f1-score & 0.97 & 0.99 & \textbf{0.95} & 0.89 & 0.98 & 0.97 & 0.99 & 0.78\\
\cline{1-2} 
\multirow{3}{*}{Ours (MP)} 
& precision & 0.99 & 0.99 & 0.95 & 1.00 & 0.95 & 0.98 & 0.98 & 0.85\\
& recall & 0.96 & 1.00 & 0.95 & 0.85 & 1.00 & 0.98 & 1.00 & 0.77\\
& f1-score & \textbf{0.97} & \textbf{1.00} & \textbf{0.95} & \textbf{0.92} & \textbf{0.98} & \textbf{0.98} & \textbf{0.99} & 0.80\\
\midrule
\midrule
Method & Metric &
person &
piano &
plant &
radio &
range\_hood &
sink &
sofa &
stairs\\
\midrule
\multirow{3}{*}{DGCNN[\citeyear{wang2019dynamic}]} 
& precision & 1.00 & 0.98 & 0.88 & 0.74 & 0.99 & 1.00 & 0.98 & 1.00\\ 
& recall & 0.95 & 0.95 & 0.84 & 0.70 & 0.97 & 0.85 & 1.00 & 0.95\\
& f1-score & \textbf{0.97} & 0.96 & \textbf{0.86} & 0.72 & 0.98& 0.92 & \textbf{0.99} & \textbf{0.97}\\
\cline{1-2} 
\multirow{3}{*}{Ours (LLE)} 
& precision & 1.00 & 1.00 & 0.86 & 0.84 & 0.99 & 0.90 & 0.98 & 1.00\\
& recall & 0.95 & 0.95 & 0.86 & 0.80 & 0.96 & 0.95 & 1.00 & 0.95\\
& f1-score & \textbf{0.97} & \textbf{0.97} & \textbf{0.86} & 0.82 & 0.97 & 0.93 & \textbf{0.99} & \textbf{0.97}\\
\cline{1-2} 
\multirow{3}{*}{Ours (MP)}
& precision & 1.00 & 0.99 & 0.86 & 0.82 & 0.98 & 1.00 & 0.98 & 1.00\\
& recall & 0.95 & 0.94 & 0.81 & 0.90 & 1.00 & 0.95 & 1.00 & 0.95\\
& f1-score & \textbf{0.97} & 0.96 & 0.84 & \textbf{0.86} & \textbf{0.99} & \textbf{0.97} & \textbf{0.99} & \textbf{0.97}\\
\midrule
\midrule
Method & Metric &
stool &
table &
tent &
toilet &
tv\_stand &
vase &
wardrobe &
xbox\\
\midrule
\multirow{3}{*}{DGCNN[\citeyear{wang2019dynamic}]} 
& precision & 0.84 & 0.84 & 0.90 & 0.99 & 0.94 & 0.83 & 0.89 & 0.94\\ 
& recall & 0.80 & 0.87 & 0.95 & 1.00 & 0.85 & 0.86 & 0.80 & 0.85\\
& f1-score & 0.82 & 0.85 & 0.93 & 1.00 & 0.89 & 0.85 & 0.84 & 0.89\\
\cline{1-2} 
\multirow{3}{*}{Ours (LLE)} 
& precision & 0.83 & 0.82 & 0.86 & 0.99 & 0.93 & 0.87 & 0.89 & 0.95\\
& recall & 0.75 & 0.83 & 0.95 & 1.00 & 0.89 & 0.89 & 0.85 & 1.00\\
& f1-score & 0.79 & 0.83 & 0.90 & 1.00 & \textbf{0.91} & \textbf{0.88} & \textbf{0.87} & \textbf{0.98}\\
\cline{1-2} 
\multirow{3}{*}{Ours (MP)} 
& precision & 0.85 & 0.88 & 0.95 & 1.00 & 0.93 & 0.81 & 0.84 & 0.77\\
& recall & 0.85 & 0.84 & 0.95 & 1.00 & 0.87 & 0.87 & 0.80 & 0.85\\
& f1-score & \textbf{0.85} & \textbf{0.86} & \textbf{0.95} & \textbf{1.00} & 0.90 & 0.84 & 0.82 & 0.81\\
\bottomrule
\end{tabular}
%}
\caption{Category-level classification result on ModelNet40 (1024 points).}
\label{tab:classification_detail}
\end{table*}

\end{document}